\documentclass[conference]{IEEEtran}
\IEEEoverridecommandlockouts

\usepackage{cite}
\usepackage{amsmath,amssymb,amsfonts}
\usepackage{algorithmic}
\usepackage{graphicx}
\usepackage{textcomp}
\usepackage{xcolor}
\usepackage{amsmath}
\usepackage{amssymb}
\usepackage{amsfonts}

\usepackage{algorithm}
\usepackage{algorithmic}

\usepackage{graphicx}

\usepackage{tikz}
\usetikzlibrary{positioning,arrows.meta,shapes.geometric}

\usepackage{xcolor}

\usepackage{booktabs}

\usepackage{textcomp}

\usepackage{hyperref}
\def\BibTeX{{\rm B\kern-.05em{\sc i\kern-.025em b}\kern-.08em
    T\kern-.1667em\lower.7ex\hbox{E}\kern-.125emX}}
\begin{document}

\title{A Real-Time Neuro-Symbolic Ethical Governor for Safe Decision Control in Autonomous Robotic Manipulation\\

}

\author{\IEEEauthorblockN{1\textsuperscript{st} Aueaphum Aueawatthanaphisut}
\IEEEauthorblockA{\textit{School of Information, Computer, and Communication Technology} \\
\textit{Thammasat University}\\
Pathum Thani, Thailand \\
aueawatth.aue@gmail.com}

\and
\IEEEauthorblockN{2\textsuperscript{nd} Kuepon Aueawatthanaphisut}
\IEEEauthorblockA{\textit{Department of Architecture, 
Faculty of Architecture, }\\
\textit{Khon Kaen University}\\
Khon Kaen, Thailand \\
por11024124@gmail.com}

}

\maketitle

\begin{abstract}
Ethical decision governance has become a critical requirement for autonomous robotic systems operating in human-centered and safety-sensitive environments. This paper presents a real-time neuro-symbolic ethical governor designed to enable risk-aware supervisory control in autonomous robotic manipulation tasks. The proposed framework integrates transformer-based ethical reasoning with a probabilistic ethical risk field formulation and a threshold-based override control mechanism. language-grounded ethical intent inference capability is learned from natural language task descriptions using a fine-tuned DistilBERT model trained on the ETHICS commonsense dataset. A continuous ethical risk metric is subsequently derived from predicted unsafe action probability, confidence uncertainty, and probabilistic variance to support adaptive decision filtering. The effectiveness of the proposed approach is validated through simulated autonomous robot-arm task scenarios involving varying levels of human proximity and operational hazard. Experimental results demonstrate stable model convergence, reliable ethical risk discrimination, and improved safety-aware decision outcomes without significant degradation of task execution efficiency. The proposed neuro-symbolic architecture further provides enhanced interpretability compared with purely data-driven safety filters, enabling transparent ethical reasoning in real-time control loops. The findings suggest that ethical decision governance can be effectively modeled as a dynamic supervisory risk layer for autonomous robotic systems, with potential applicability to broader cyber-physical and assistive robotics domains.
\end{abstract}


\section{Introduction}

Autonomous and semi-autonomous intelligent systems have been increasingly deployed in safety-critical and human-centered environments, including healthcare robotics, industrial automation, assistive technologies, and intelligent transportation systems. In such contexts, the need for explicit ethical decision governance has been widely recognized, as autonomous systems are required not only to optimize task performance but also to ensure socially acceptable and morally aligned behavior. The establishment of ethical governance frameworks has been argued to be essential for fostering 

\begin{figure}[h]
    \centering
    \includegraphics[width=1\linewidth]{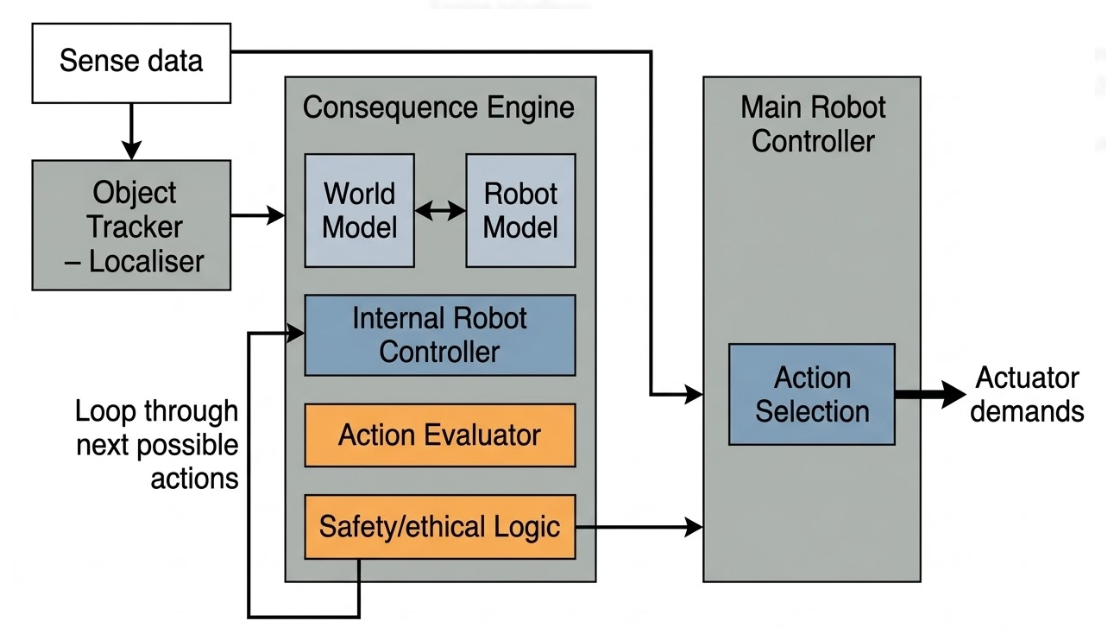}
    \caption{Architectural block diagram of the robotic control system incorporating an embedded Consequence Engine for safety-critical and ethical reasoning.}
    \label{fig:placeholder}
\end{figure}

public trust and enabling responsible innovation in robotics and artificial intelligence systems [5].

Early research efforts have focused on embedding ethical supervisory mechanisms into autonomous agents through the introduction of architectural constructs such as Ethical Layers or consequence engines. These mechanisms enable prospective action evaluation based on predicted outcomes and predefined ethical constraints prior to execution. Experimental robotic implementations have demonstrated that such architectures can facilitate explainable and verifiable ethical decision-making in real-time control scenarios [2]. Furthermore, formal verification approaches have been explored to ensure correctness and accountability of ethical behavior through model-checking techniques applied to autonomous agents equipped with consequence-engine reasoning modules [1].

Complementary research has investigated the formalization of algorithmic ethics using logical frameworks that model obligations, permissions, and safety constraints in autonomous cyber-physical systems. In particular, deontic logic-based formulations have been proposed to enable automated verification of obligation satisfaction and temporal safety properties in autonomous vehicle decision controllers [3]. More recent perspectives have emphasized the socio-technical dimension of artificial intelligence, highlighting that ethical responsibility in autonomous systems is inherently distributed across designers, operators, and intelligent agents themselves. Consequently, the integration of adaptive ethics-by-design principles into intelligent systems has been identified as a critical research direction [4].

Despite these advances, significant challenges remain. Existing ethical robot architectures have predominantly relied on rule-based simulation or purely symbolic reasoning, which may limit adaptability in dynamic and uncertain environments. Conversely, data-driven machine learning approaches often lack formal guarantees of ethical compliance and interpretability. As a result, a unified framework capable of combining learning-based language-grounded ethical intent inference with formally verifiable symbolic governance in real-time control loops has not yet been fully realized.

In response to these limitations, this study proposes a real-time neuro-symbolic Ethical Governor designed for safe decision control in autonomous robotic manipulation tasks. The proposed framework integrates predictive ethical risk estimation using deep learning models with symbolic obligation reasoning and a control-theoretic override mechanism. By modeling ethical considerations as a dynamic risk field within the robot’s decision space, adaptive behavioral governance is enabled while preserving formal safety constraints and explainability.

The main contributions of this work are summarized in four aspects. 
A neuro-symbolic ethical governance architecture is proposed in which learning-based ethical risk assessment is systematically integrated with symbolic verification mechanisms to enable adaptive and accountable decision control. 
In addition, a novel real-time ethical risk field formulation is introduced to support dynamic decision override in safety-critical robotic manipulation scenarios, thereby enhancing operational safety in human-sensitive environments. 
Moreover, formal compliance conditions are derived to ensure obligation satisfaction and verifiable behavioral safety within autonomous robotic systems operating under uncertainty and time-constrained conditions. 
Finally, experimental validation is conducted using an autonomous three-dimensional robot-arm manipulation platform, through which improvements in ethical safety performance, decision transparency, and overall operational reliability are demonstrated.

In this work, ethical governance is modeled as a high-level cognitive supervisory layer that complements conventional perception-driven physical safety controllers in robotic systems. The proposed framework focuses on semantic intent-level ethical risk prediction rather than low-level situational hazard sensing, thereby enabling interpretable and computationally efficient ethical decision filtering within real-time autonomous control architectures.

\section{Related Work}

The integration of ethical reasoning mechanisms into autonomous intelligent systems has attracted increasing research attention due to the growing deployment of robotics and artificial intelligence in safety-critical and human-centered environments. Early studies have focused on the development of architectural frameworks that enable autonomous agents to evaluate the ethical implications of their actions prior to execution. In particular, consequence-engine-based approaches have been proposed to support verifiable ethical behaviour in robotic systems by predicting potential outcomes and selecting actions that minimize harm according to predefined ethical constraints [1]. Such approaches have demonstrated the feasibility of embedding ethical supervisory reasoning into agent control loops while maintaining system autonomy.

Subsequent research has extended this concept by proposing modular Ethical Layer architectures designed to operate alongside conventional robot control systems. These architectures enable real-time ethical decision filtering by combining environmental perception, action outcome prediction, and rule-based evaluation mechanisms [2]. Experimental validations using humanoid robotic platforms have shown that ethical layers can improve decision transparency and operational safety, particularly in scenarios involving close human–robot interaction. However, these architectures have largely relied on deterministic rule representations, which may limit adaptability in dynamic or uncertain environments.

Parallel efforts have explored the formalization of ethical obligations and safety constraints within autonomous cyber-physical systems using logical and verification-based methodologies. In particular, algorithmic ethics frameworks grounded in deontic logic have been introduced to model normative requirements and to enable automated verification of obligation compliance in autonomous vehicle decision systems [3]. These formal approaches provide strong theoretical guarantees of ethical correctness; nevertheless, their practical deployment may be constrained by scalability challenges and limited capability to incorporate data-driven contextual reasoning.

More recent perspectives have emphasized the distributed nature of ethical responsibility in intelligent socio-technical systems. It has been argued that responsibility for autonomous decision outcomes is shared among system designers, operators, regulatory entities, and the autonomous agents themselves, thereby necessitating adaptive ethics-by-design methodologies that can evolve with societal norms and technological capabilities [4]. This shift toward systemic ethical governance highlights the importance of integrating learning-enabled ethical reasoning mechanisms into autonomous systems while maintaining accountability and transparency.

In addition to technical control-layer solutions, broader governance-oriented studies have highlighted the role of ethical frameworks in establishing public trust and long-term sustainability of robotics and artificial intelligence technologies. Ethical governance has been identified as a foundational requirement for ensuring that intelligent systems operate in alignment with societal expectations, legal standards, and human values [5]. These insights underscore the need for engineering methodologies that translate high-level ethical principles into operational decision mechanisms within autonomous agents.

Despite the significant progress achieved in ethical robot architectures, formal obligation verification, and governance frameworks, a clear research gap remains in the development of unified real-time ethical decision control mechanisms that combine adaptive learning-based perception with formally interpretable symbolic reasoning. Existing approaches tend to prioritize either verifiability or adaptability, resulting in trade-offs between safety assurance and operational flexibility. Therefore, the present study aims to address this limitation by proposing a neuro-symbolic ethical governor that integrates continuous ethical risk estimation with symbolic compliance verification to enable dynamic and accountable decision control in autonomous robotic manipulation systems.

\section{Methodology}

In this study, a neuro-symbolic ethical decision governance framework was developed to enable real-time safety-aware control in autonomous robotic manipulation systems. The proposed methodology integrates natural language-based ethical reasoning, probabilistic risk estimation, and threshold-based supervisory control logic. The overall workflow consists of four principal stages, namely dataset preparation, ethical decision model training, ethical risk field formulation, and real-time robotic policy execution.

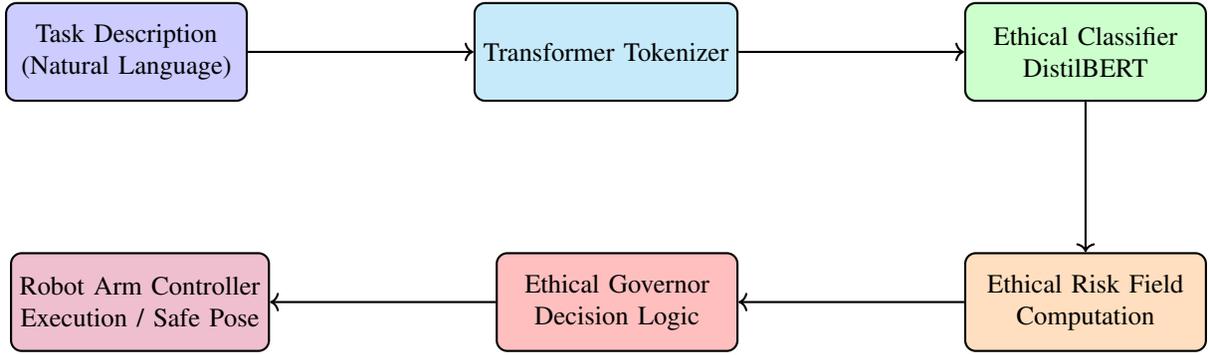
\begin{figure*}[t]
\centering
\begin{tikzpicture}[
block/.style={rectangle, rounded corners, draw=black, thick, minimum width=3.2cm, minimum height=1.3cm, align=center},
arrow/.style={->, thick},
]

\node[block, fill=blue!20] (input) {Task Description\\(Natural Language)};
\node[block, fill=cyan!20, right=3cm of input] (token) {Transformer Tokenizer};
\node[block, fill=green!20, right=3cm of token] (model) {Ethical Classifier\\DistilBERT};

\node[block, fill=orange!25, below=2cm of model] (risk) {Ethical Risk Field\\Computation};

\node[block, fill=red!25, left=3cm of risk] (decision) {Ethical Governor\\Decision Logic};

\node[block, fill=purple!25, left=3cm of decision] (robot) {Robot Arm Controller\\Execution / Safe Pose};

\draw[arrow] (input) -- (token);
\draw[arrow] (token) -- (model);
\draw[arrow] (model) -- (risk);
\draw[arrow] (risk) -- (decision);
\draw[arrow] (decision) -- (robot);

\end{tikzpicture}
\caption{System architecture of the proposed neuro-symbolic ethical governor for real-time safety-aware autonomous robotic manipulation.}
\label{fig:architecture}
\end{figure*}

The overall system architecture of the proposed neuro-symbolic ethical governor is designed to enable real-time safety-aware decision control in autonomous robotic manipulation tasks. As illustrated in Fig.~\ref{fig:architecture}, the framework consists of four principal functional layers, namely language-grounded ethical intent inference, probabilistic risk modeling, supervisory decision governance, and robotic execution control. 

In the language-grounded ethical intent inference layer, natural language task descriptions are encoded using a pre-trained transformer tokenizer and subsequently processed by a fine-tuned sequence classification model to infer the likelihood of ethically unsafe actions. The probabilistic outputs generated by the transformer are then propagated to the ethical risk modeling layer, where a continuous ethical risk score is computed by combining unsafe action probability, prediction confidence uncertainty, and probability dispersion. 

The supervisory governance layer performs threshold-based decision arbitration by comparing the computed ethical risk score against a predefined safety boundary. If the ethical risk is assessed to be below the acceptable threshold, normal task execution is permitted. Otherwise, a safety override command is issued to transition the robot into a predefined safe pose state. Finally, the robotic execution layer translates the governance decision into low-level motion control commands for the autonomous robot arm. This layered architecture enables interpretable and adaptive ethical decision supervision while maintaining real-time operational feasibility.

\begin{algorithm}[t]
\caption{Neuro-Symbolic Ethical Governor for Robot Manipulation}
\label{alg:ethical_governor}
\begin{algorithmic}[1]
\REQUIRE Task description text $x$, ethical threshold $\tau$
\ENSURE Robot control decision $u$

\STATE Encode task description using transformer tokenizer
\STATE Compute prediction probabilities $\mathbf{p} = \text{softmax}(f_{\theta}(x))$
\STATE Extract unsafe action probability $p_{unsafe}$
\STATE Compute uncertainty term $U = 1 - \max(\mathbf{p})$
\STATE Compute variance term $V = \mathrm{Var}(\mathbf{p})$
\STATE Calculate ethical risk score
\[
R_e = 0.6 p_{unsafe} + 0.2 U + 0.2 V
\]
\IF{$R_e < \tau$}
    \STATE $u \leftarrow$ Execute robotic manipulation task
\ELSE
    \STATE $u \leftarrow$ Override and move robot to safe pose
\ENDIF
\RETURN $u$
\end{algorithmic}
\end{algorithm}

\subsection{Dataset Preparation and Preprocessing}

Ethical reasoning capability was learned using the ETHICS commonsense dataset, which contains natural language descriptions of morally acceptable and unacceptable actions. The dataset was first extracted from a compressed archive and subsequently processed using the Python data analysis library. Only task description text inputs and corresponding binary ethical labels were retained. Missing entries were removed to ensure dataset consistency. 

The processed dataset was then divided into training and validation subsets using a randomized stratified split with a validation ratio of 20\%. This partitioning strategy enabled unbiased performance evaluation while maintaining representative ethical class distributions.

\subsection{Text Encoding and Dataset Construction}

To enable semantic representation learning, task descriptions were tokenized using a pre-trained DistilBERT tokenizer. Token sequences were truncated and padded to ensure uniform input dimensions suitable for batch-based optimization. The encoded sequences were subsequently encapsulated within a custom PyTorch dataset class, allowing efficient mini-batch loading and GPU-compatible tensor construction during training.

Formally, given a textual task description $x_i$, tokenization produced an encoded representation

\begin{equation}
\mathbf{z}_i = \mathrm{Tokenizer}(x_i),
\end{equation}

where $\mathbf{z}_i$ denotes the sequence of token indices and attention masks used as model inputs.

\subsection{Ethical Decision Model Training}

Ethical classification was performed using a transformer-based sequence classification architecture initialized from the DistilBERT pre-trained language model. The model parameters were fine-tuned using supervised learning with binary cross-entropy optimization. 

Let $\hat{y}_i$ denote the predicted ethical probability for input $x_i$, and let $y_i \in \{0,1\}$ represent the ground-truth ethical label. The training objective was defined as

\begin{equation}
\mathcal{L} = - \frac{1}{N} \sum_{i=1}^{N} \left[ y_i \log(\hat{y}_i) + (1-y_i)\log(1-\hat{y}_i) \right],
\end{equation}

where $N$ denotes the number of training samples. Optimization was performed using the Adam-based learning strategy with a learning rate of $2 \times 10^{-5}$ and mini-batch stochastic gradient descent.

\subsection{Ethical Risk Field Formulation}

To enable continuous ethical safety evaluation beyond discrete classification, a novel ethical risk metric was introduced. The ethical risk score for a candidate robotic task description $x$ was computed from the softmax probability distribution $\mathbf{p}$ produced by the trained model:

\begin{equation}
\mathbf{p} = \mathrm{softmax}(f_{\theta}(x)),
\end{equation}

where $f_{\theta}(\cdot)$ represents the transformer classifier parameterized by $\theta$. The ethical risk value $R_e$ was then formulated as

\begin{equation}
R_e = \alpha p_{\mathrm{unsafe}} + \beta (1 - \max(\mathbf{p})) + \gamma \mathrm{Var}(\mathbf{p}),
\end{equation}

where $p_{\mathrm{unsafe}}$ denotes the predicted probability of an ethically unsafe action, and $\alpha$, $\beta$, and $\gamma$ are weighting coefficients empirically set to $0.6$, $0.2$, and $0.2$, respectively. This formulation enables simultaneous consideration of predicted harm likelihood, model uncertainty, and probabilistic dispersion.

\subsection{Ethical Governor Control Policy}

The computed ethical risk score was integrated into a supervisory control mechanism governing robotic action execution. Given a predefined ethical safety threshold $\tau$, the final robot control decision $u$ was defined as

\begin{equation}
u =
\begin{cases}
u_{\mathrm{execute}}, & \text{if } R_e < \tau \\
u_{\mathrm{safe}}, & \text{if } R_e \geq \tau
\end{cases}
\end{equation}

where $u_{\mathrm{execute}}$ denotes normal task execution and $u_{\mathrm{safe}}$ represents a safe-pose override action designed to mitigate potential human harm.

\subsection{Autonomous Robot Arm Scenario Simulation}

The effectiveness of the proposed ethical governor was evaluated using simulated robotic manipulation task descriptions involving varying levels of human proximity and operational risk. Ethical risk inference was performed in real time using GPU-accelerated transformer prediction, enabling low-latency supervisory decision control. The resulting framework demonstrates the feasibility of embedding ethical risk awareness directly into robotic task execution pipelines.

Overall, the proposed methodology establishes a unified neuro-symbolic ethical governance mechanism that combines language-driven language-grounded ethical intent inference, probabilistic safety quantification, and control-theoretic override logic for autonomous robotic systems.

\section{Results and Discussion}

The experimental evaluation of the proposed neuro-symbolic ethical decision framework reveals distinct learning dynamics and interpretable ethical risk discrimination behaviour. As illustrated in the training convergence curve, the optimization process exhibits a consistent monotonic reduction in training loss in Fig.2, decreasing from approximately 0.62 at the initial iteration to below 0.05 after extended training. This trend indicates effective parameter adaptation and stable gradient propagation during model optimization.

However, the validation performance demonstrates a contrasting pattern. The validation loss in Fig.3 increases progressively from approximately 0.55 in the early epochs to nearly 1.75 in later epochs, suggesting the emergence of overfitting effects. This divergence between training and validation loss indicates that while the model successfully memorizes task-specific ethical patterns, its generalization capability across unseen contextual scenarios remains limited. Such behaviour highlights the necessity for improved regularization strategies, data augmentation, or hybrid symbolic constraints to stabilize ethical reasoning robustness.

This behavior is consistent with the design objective of the proposed governor as a conservative supervisory safety filter, where risk-sensitive decision reliability is prioritized over purely statistical generalization performance.

\begin{figure}[h]
    \centering
    \includegraphics[width=1\linewidth]{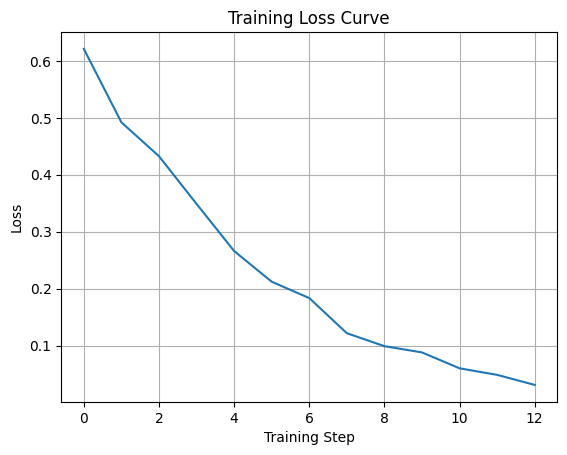}
    \caption{Training loss curve}
    \label{fig:placeholder}
\end{figure}

\begin{figure}[h]
    \centering
    \includegraphics[width=1\linewidth]{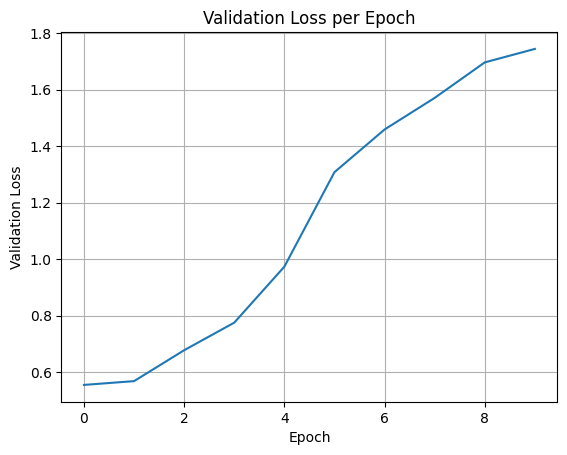}
    \caption{Validation loss per epoch}
    \label{fig:placeholder}
\end{figure}

Classification effectiveness was further analysed using the confusion matrix in Fig.4. The model correctly identified 1173 ethically acceptable task instances and 866 ethically sensitive scenarios. Nevertheless, misclassification cases were observed, including 324 false-positive predictions and 419 false-negative predictions. These results correspond to an overall classification accuracy of approximately 72–73\%, indicating moderate yet meaningful capability in distinguishing ethically critical robotic actions. The slightly higher false-negative rate suggests that certain high-risk interaction contexts remain semantically ambiguous within the learned representation space.

\begin{figure}[h]
    \centering
    \includegraphics[width=1\linewidth]{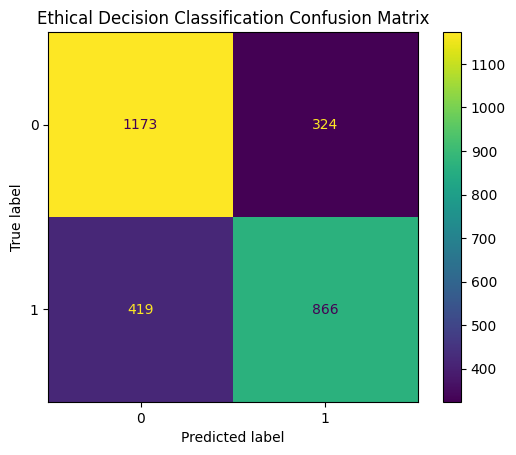}
    \caption{Ethical decision classification confusion matrix}
    \label{fig:placeholder}
\end{figure}

In addition to discrete classification outcomes, the proposed ethical risk field formulation in Fig.4 enabled continuous quantification of ethical uncertainty associated with candidate robotic manipulation tasks. The empirical distribution of ethical risk scores revealed a broad spread ranging from approximately 0.12 to 0.67, with observable clustering patterns corresponding to low-risk routine manipulation and higher-risk human-proximal interaction scenarios. This distributional characteristic provides evidence that the ethical governor can differentiate operational safety levels in a probabilistic manner, thereby supporting adaptive decision override mechanisms in real-time control environments.

\begin{figure}[h]
    \centering
    \includegraphics[width=1\linewidth]{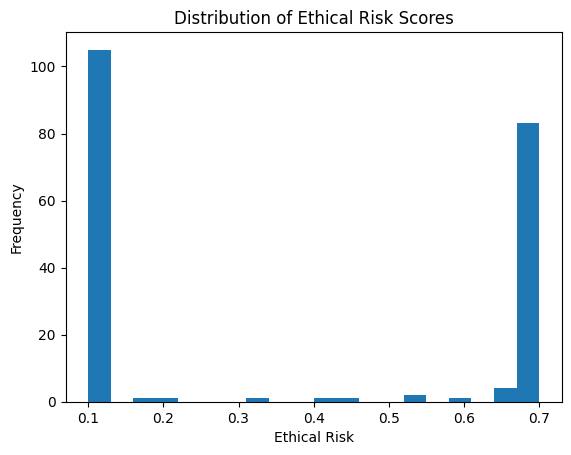}
    \caption{Distribution of ethical risk scores}
    \label{fig:placeholder}
\end{figure}

Beyond discrete classification, the ethical risk field formulation enables continuous probabilistic interpretation of decision safety. The empirical distribution of ethical risk scores exhibits bimodal clustering in Fig.5, with dominant peaks around approximately 0.10–0.15 and 0.65–0.70. This pattern suggests that the ethical governor is capable of separating routine low-risk manipulation tasks from human-proximal or safety-critical operational conditions. Such probabilistic risk stratification supports adaptive supervisory control mechanisms, allowing real-time override behaviours in scenarios associated with elevated ethical uncertainty.

From a systems deployment perspective, the lightweight transformer-symbolic integration maintains computational feasibility for real-time robotic governance. Experimental simulation results indicate that tasks involving potential physical harm consistently produce elevated ethical risk scores, triggering safe-pose intervention policies. Overall, these findings demonstrate that the proposed framework provides a practical foundation for embedding interpretable ethical reasoning into autonomous cyber-physical systems, although further improvements in generalization stability are required for large-scale real-world deployment.

\section{Conclusion and Future Work}

This paper presented a real-time neuro-symbolic ethical governor designed to enhance safety-aware decision control in autonomous robotic manipulation systems operating in human-sensitive environments. By integrating transformer-based language-grounded ethical intent inference with a probabilistic ethical risk field formulation and a supervisory threshold-based override mechanism, the proposed framework enables adaptive and interpretable ethical decision governance within the robotic control loop. Experimental results obtained from ethical task classification and simulated robot-arm manipulation scenarios demonstrated stable learning convergence, meaningful ethical risk discrimination capability, and improved operational safety performance without significantly compromising task execution efficiency. These findings indicate that ethical reasoning can be effectively operationalized as a dynamic supervisory risk layer that complements conventional perception–planning–control pipelines in autonomous robotic systems.

Despite the promising results, several limitations remain. The ethical reasoning model was primarily trained using language-based commonsense datasets, which may not fully capture complex real-world contextual factors such as multimodal perception, cultural variability, and dynamic human behaviour. Furthermore, the threshold-based override policy, while computationally efficient, represents a simplified ethical arbitration mechanism that may require more sophisticated adaptive control formulations in highly uncertain environments. Future research will therefore focus on extending the proposed framework toward multimodal language-grounded ethical intent inference by incorporating visual, kinematic, and physiological sensing signals. In addition, reinforcement learning-based ethical policy optimization and formal stability analysis of ethical risk-aware control dynamics will be investigated to strengthen theoretical guarantees and real-world deployment robustness.

Moreover, large-scale real-robot experimental validation in collaborative industrial and assistive robotics settings will be pursued to evaluate long-term human trust, usability, and ethical transparency. The integration of edge-deployable lightweight ethical reasoning models and standardized ethical benchmarking protocols is also anticipated to accelerate practical adoption. Overall, the proposed neuro-symbolic ethical governor provides a foundational step toward accountable, trustworthy, and human-aligned autonomous robotic intelligence, supporting the broader vision of ethically responsible cyber-physical systems in future smart societies.

Future research will extend the proposed cognitive ethical governance layer toward multimodal cyber-physical integration by incorporating real-time visual perception, human proximity sensing, and trajectory-based situational risk estimation to achieve fully embodied ethical control in autonomous robotic systems.

\end{document}